\documentclass[conference]{IEEEtran}
\IEEEoverridecommandlockouts

\usepackage{cite}
\usepackage{amsmath,amssymb,amsfonts}
\usepackage{algorithmic}
\usepackage{graphicx}
\usepackage{textcomp}
\usepackage{listings}
\usepackage{xurl}
\usepackage{comment}
\usepackage{xcolor}

\begin{document}

\title{An Accurate and Low-Parameter Machine Learning Architecture for Next Location Prediction}

\makeatletter
\newcommand{\linebreakand}{%
  \end{@IEEEauthorhalign}
  \hfill\mbox{}\par
  \mbox{}\hfill\begin{@IEEEauthorhalign}
}
\makeatother
\author{\IEEEauthorblockN{Calvin Jary} 
\IEEEauthorblockA{\textit{Systems and Computer Engineering} \\
\textit{Carleton University, Canada}\\
calvinjary@sce.carleton.ca}

\and
\IEEEauthorblockN{Nafiseh Kahani}
\IEEEauthorblockA{\textit{Systems and Computer Engineering} \\
\textit{Carleton University, Canada}\\
kahani@sce.carleton.ca}
}

\maketitle

\begin{abstract}


Next location prediction is a discipline that involves predicting a user’s next location. Its applications include resource allocation, quality of service, energy efficiency, and traffic management. This paper proposes an energy-efficient, small, and low parameter machine learning (ML) architecture for accurate next location prediction, deployable on modest base stations and edge devices.
To accomplish this we ran a hundred hyperparameter experiments on the full human mobility patterns of an entire city, to determine an exact ML architecture that reached a plateau of accuracy with the least amount of model parameters. We successfully achieved a reduction in the number of model parameters within published ML architectures from $202$ million down to $2$ million. This reduced the total size of the model parameters from $791$ MB down to $8$ MB. Additionally, this decreased the training time by a factor of four, the amount of graphics processing unit
(GPU) memory needed for training by a factor of twenty, and the overall accuracy was increased from $80.16\%$ to $82.54\%$. This improvement allows for modest base stations and edge devices
which do not have a large amount of memory or storage, to deploy and utilize the proposed ML architecture for next location prediction.


\end{abstract}


\newcommand{\para}[2]{\textit{\textbf{#1}: {#2} \\}}
\newcommand{\codee}[1]{\small\textit{#1}\normalsize}
\newcommand{\code}[1]{\small\texttt{#1}\normalsize}

\section{Introduction}
\label{sec:introdution}

The ability to predict a user's next location has numerous practical applications, spanning a diverse range of fields such as personalized recommendations, efficient traffic management ~\cite{besse2017destination}, optimized energy consumption \cite{energyconsumption}, and network resource optimization~\cite{nextlocationsurvey,ying2011semantic}. More specifically, in the context of mobile networks (e.g., 5G) which is the context of this work, next location prediction enables operators to reduce network burden, optimize network resources and mobility management~\cite{energyconsumption,yan2019pecs,yan2019assessing}.

Predictive models have been widely used to anticipate a user's next location based on their location history, but these models face significant challenges in terms of accuracy and complexity. While numerous studies have attempted to improve upon these models, a recent survey~\cite{nextlocationsurvey} highlights that there is still considerable work to be done in developing accurate and optimized machine learning architectures. In the context of mobile networks where predictive models may need to be deployed in base stations and edge devices with limited memory and computational capability, there are advantages to both improvements in model accuracy as well as model simplicity. One may even need to sacrifice some amount of accuracy in order to have a model with fewer parameters that can be easily deployed on a base station or an edge device. 

Finding the optimal balance between accuracy and simplicity is therefore a crucial consideration in developing predictive models for users' future locations in mobile networks context. Thus, in this research we conduct an empirical study to attempt an optimization of both accuracy and complexity of predictive models for next location prediction. Our focus is on recurrent neural network (RNN) architectures and a mobility dataset collected from Changchun, China \cite{Du2018}.

The advantage of our ML architecture is in its ability to train on even a very modest GPU within a few hours on a city-wide dataset, and achieve an approximate $82$\% accuracy in predicting the next location of any individual user. This prediction is based solely on the user's current location along with their previous four locations. Importantly, our model is designed to work effectively on any location, with any user, and on any day of the week.

Our work makes the following contributions towards an accurate, small and deployable machine learning (ML) architecture for wireless mobility predication:

 \begin{itemize}
     \item  We evaluated how different model hyperparameters would need to scale up and also scale down to efficiently address a larger city or a smaller city. With coorespondingly more users, more locations and more data or with fewer users, fewer locations and less data. 
     \item We explored how the amount of training in terms of both epochs and iterations would change with more data as well as less data in order to fully train the model given our proposed architecture. This was done to have a more complete ML solution given diverse datasets and applications.
    \item We developed a model that has very few parameters and therefore could easily be deployed and maintained, as well as be able to operate on modest base stations and edge devices.
 \end{itemize}
 
The remainder of the paper is structured as follows. Section~\ref{sec:background} discusses the background and related works. Section~\ref{sec:methods} describes our approach, followed by Section~\ref{sec:validation}, which discusses our validation. We present our results and discussions in Section~\ref{sec:results-discussion}. Finally, Section~\ref{sec:conclusion} concludes the paper.

\section{Background and Related Works}
\label{sec:background}

Human mobility prediction has become increasingly viable as a discipline due to the growth in user location information in the form of massive trajectory data sets, as well as improvements made to the field of ML and ever increasing compute performance. Motion behaviors can be passively collected by mobile phones in terms of Bluetooth data, GPS data or WiFi data. Various ML approaches can then be used to calculate what the future location of a user may be, which has given rise to next location prediction.

Next location prediction has been effectively surveyed, and has been addressed by multiple groups. Solutions began with statistical methods such as semantic trajectory mining \cite{ying2011semantic} and Markov models \cite{gambs2012next, chen2014nlpmm}, solutions then improved with recurrent neural networks (RNNs) \cite{liu2016predicting, di2017semantic} and long short-term memory networks (LSTMs) \cite{Crivellari2020, BDLSTM2018} and have continued most recently with attention networks and transformers \cite{seq2seq2018, trajectory2019, selfattention2021, finalref1}. 
Furthermore, these modern ML architectures are able to learn individual user mobility patterns without any manual feature generation, feature extraction, or any additional context which historically was critical to model accuracy \cite{feature2018}. The current location of the user, along with a recent history of their previous locations is all that is required. This has great advantages in both the training and the deployment of the ML systems for this application \cite{trafficflow2022, TRN2018}.

In the recent work based on the LSTMs \cite{Crivellari2020, BDLSTM2018, di2017semantic}, learning individual user mobility patterns is accomplished by each individual location being uniquely represented by an embedding vector, which is a low-dimensional dense vector that provides the features that are fed into and utilized by the RNN block. This dense vector is of a pre-defined length and randomly initialized. Through training via back-propagation, the values are updated to maximize their utility in accurately predicting the next location. Essentially as training progresses, the embedding values for each location adopt meaningful values, whereby locations co-occurring in the same user movement patterns take on similar values for one or more embedding dimensions \cite{finalref2}.

We selected the approach proposed by Crivellari and Beinat \cite{Crivellari2020} as our baseline. The reason being this was already a simple and elegant ML architecture for next location prediction. This architecture did not require dataset-specific features such as latitude and longitude for spacial context, it did not require convolutional layers as a means of contextualizing neighbouring locations, or the use of attention networks which are high in parameter and computational cost for inference. In comparing our model to this previous work, we aim to reduce the number of model parameters as much as possible, while maintaining a high accuracy.

\section{Approach}
\label{sec:methods}

Our approach involves the development of a predictive model with the main objective of identifying the most likely next movement pattern of a user based on their location history. We aim to maintain a simple architecture, minimizing the number of model parameters, while simultaneously maintaining a high level of accuracy. 

The model consists of three main components: an embedding layer, a block of RNN layers, and a linear layer with a softmax function. To encode input trajectories, we associate each location identifier with a corresponding embedding vector. This enables the encoding of input trajectories into sequences of embeddings, which capture temporal and location dependencies, and which are subsequently fed into an RNN block. The RNN block consists of stacked RNN layers that are fed and compute the location sequence. This is finally fed into a fully connected linear layer which computes the final trajectory representation. A softmax function is then applied to generate a probabilistic distribution. Figure~\ref{figure-architecture} shows our high-level model architecture.

In this section, we provide a detailed description and rationale for having each component of our model, as well as an overview of how these components integrate into the overall architecture.

\begin{figure}[t]  
  \centering
  \includegraphics[width=0.45\textwidth]{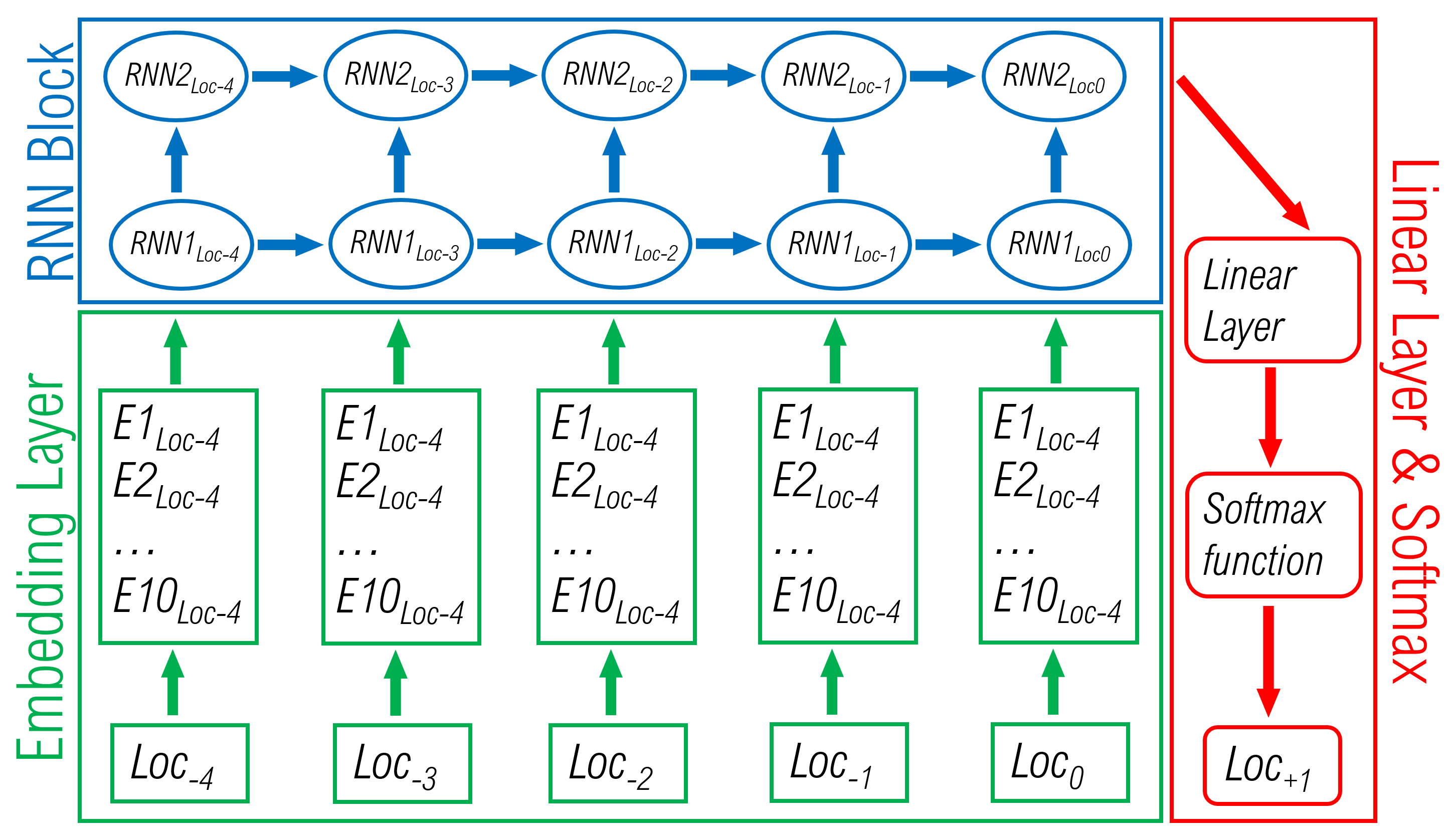}
  \caption{High-level architecture of the proposed ML model}
  \label{figure-architecture}
\end{figure}

\subsection{Input Size}
The number of inputs, also known as the window size, represents the total number of locations provided to the model, including the current location. A larger window size allows the model to consider additional previous locations, which provides more context for the model to make accurate predictions. For example, if the window size is set to five, it means that the model uses the current location along with the four most recently visited locations, to calculate the subsequent location. The window size is a hyperparameter that requires a trade-off between model accuracy, and the need for a more extensive history of past locations visited by the user. We will discuss the details of hyperparameter tuning for the input size in Section \ref{sec:results-discussion}.\\

\subsection{Embedding Dimensions}
As shown in Figure~\ref{figure-architecture}, each location is represented by an embedding, which is a dense vector consisting of a 1D array of numbers. 
In our application, these embeddings represent the correlation between different locations. For example, if it is common for users to travel from location \textit{A} to location \textit{B}, the embeddings for locations \textit{A} and \textit{B} will have similar values for one or more elements of the embedding vector. The specific values themselves are not important. What is important is that they are similar, indicating a relationship between the two locations. Entirely unrelated locations would have different values for those same embedding elements, reflecting their own encoding relationship. The embedding hyperparameter simply determines the dimensionality of this embedding vector, which influences the model's ability to capture and differentiate location relationships. We will discuss the details of hyperparameter tuning for the embedding dimensions in Section \ref{sec:results-discussion}.

\subsection{RNN Block}
Within the application of ML for next location prediction, LSTMs have been widely employed in the literature \cite{Crivellari2020, BDLSTM2018}. This is due to being a relatively sophisticated and modern ML architecture. However, in order to minimize the complexity of our model, while insuring high accuracy, the simpler RNN may prove to be superior. This is due to the fact that LSTMs were specifically designed to address the computational issues that arise from very long inputs, which are not an issue within the context of mobility prediction, which has a small number of inputs \cite{LSTMbackground2021}. Therefore, to optimize model simplicity and accuracy, we examined the benefits of both RNNs and LSTMs in next location prediction. Both architectures consist of interconnected neural networks that process input sequences step by step, computing their hidden state until the end of the sequence is reached. The output of the last step represents the comprehensive vector characterization of the sequence, which is then used for prediction. In our model we have used multiple RNNs layers, where each layer receives the output of the previous layer. In Section \ref{sec:results-discussion} we will show that by leveraging the RNN block, our model can effectively process sequential data and capture relevant dependencies for accurate trajectory prediction.

\subsection{Hidden Nodes}  
The number of hidden nodes per RNN cell is another critical hyperparameter that directly impacts the model's complexity and its ability to process information. With our dataset, our aim is to predict every user's next location out of the $2134$ possible locations in the city (this is the number of unique classes within our dataset). 

To provide context for the appropriate number of hidden nodes, we consider the scale of the problem in terms of the large number of classes. We express this relationship using the metric ``\% of hidden nodes to num classes''. For example, a value of $10\%$, represents $10\%$ as many hidden nodes as the total number of classes, resulting in $213$ hidden nodes for the $2134$ total classes (total locations). This approach provides the necessary context for determining the optimal number of hidden nodes in relation to the complexity of the classification task, and it makes our work more transferable to other datasets and cities.

\subsection{Number of Layers}
The number of hidden layers in a model has a profound impact on its parameter count and capacity. Each additional layer of hidden nodes significantly enhances the model's abstraction, information processing, and representational capacity. While one layer is typical for straightforward applications, deep learning typically starts with two or more layers. Crivellari's approach has two layers \cite{Crivellari2020}, suggesting that mobility prediction can be considered a relatively simple deep learning problem. In our approach, we investigated the benefits to model accuracy and the increase in model parameters \cite{reviewersuggestion1} from using different numbers of layers. In our experiments, we also found two layers to be the ideal number.

\subsection{Linear Layer and Softmax Function}
Our model provides accurate location predictions through the integration of a linear layer on top of the RNN block. The purpose of this fully-connected linear layer is to compute the mobility likelihood scores for every possible location. This is followed by a softmax function, which generates a probability distribution based on the final trajectory vector characterization as an input, summing to a value of $1$. This represents the likelihood that each potential location will be the next location reached by the user. This distribution provides valuable insights into the user's next possible move, enabling informed decision-making. Figure~\ref{figure-architecture} explains this process in detail. It shows how information is efficiently flowed within the model, highlighting the importance of both the linear layer and the softmax function in accurately predicting future locations.

\section{Validation}
\label{sec:validation}

 This section reports on the experiments we conducted to assess the impact of our proposed model on the effectiveness and cost of next location prediction. We first discuss and motivate two research questions. We then describe the dataset of the study, and explain our experimental process. Finally, we discuss the results and answer the RQs.

\subsection{Research Questions}

\textit{RQ1. A simpler ML architecture in the context of next location prediction.}

\begin{itemize}
    \item \textbf{RQ1.1} Is it possible to create a simpler (fewer parameter) model that has equal or greater accuracy to existing models? This is desirable as it allows edge devices limited by power, compute performance and memory to execute the proposed model.
    
    \item \textbf{RQ1.2}: What are all the benefits of a simpler model? This can include fewer parameters, a smaller amount of space needed to store the model, decreased training time, the ability to be trained on less powerful hardware, and deployable on lower-end edge devices.

\end{itemize}

\textit{RQ2. The impact of scaling the model with different hyperparameter values.}

\begin{itemize}
    \item \textbf{RQ2.1} How does varying the window size impact the accuracy and number of parameters of our ML model architecture, within the application of next location prediction?
    \item \textbf{RQ2.2} How does varying the number of hidden nodes impact the accuracy and number of parameters of our ML model architecture, within the application of next location prediction?
    \item \textbf{RQ2.3} How does varying the number of embedding dimensions impact the accuracy and number of parameters of our ML model architecture, within the application of next location prediction?
\end{itemize}

\subsection{Dataset}
We have used an open source, city-wide dataset from Changchun Municipality, China~\cite{Du2018}. This dataset features $2,066,000$ anonymous users and their mobility history over the course of seven days. Each user's trajectory has been recorded at a temporal resolution of one hour, starting at 2:00 am and continuing until 2:00 am the next day. This may also be viewed as a $24$-dimensional location vector for $24$ consecutive hours.

There are $7251$ cellular base stations within the city; however, a single location may have multiple base stations for the purposes of servicing high demand. As a result, the $7251$ cellular base stations have been clustered into $2134$ discrete locations. This number refers to the number of unique classes within the dataset we aim to predict accurately. Furthermore, when multiple users share the exact same $24$-hour location journey (i.e., the same $24$-dimensional location vector) they are clustered within the same group, essentially as a form of data compression. Figure~\ref{figure-dataintro} provides a numerical example of our dataset. In this example, $63$ users all share an identical $24$-hour period, starting with location $11528$ at 2:00 am. Therefore, the $2,066,000$ anyonymous users have been clustered into $197,415$ unique $24$-hour location paths across the entire seven days, after removing all $24$-hour location paths that have any missing data.

\begin{figure}[t]  
  \centering
  \includegraphics[width=0.45\textwidth]{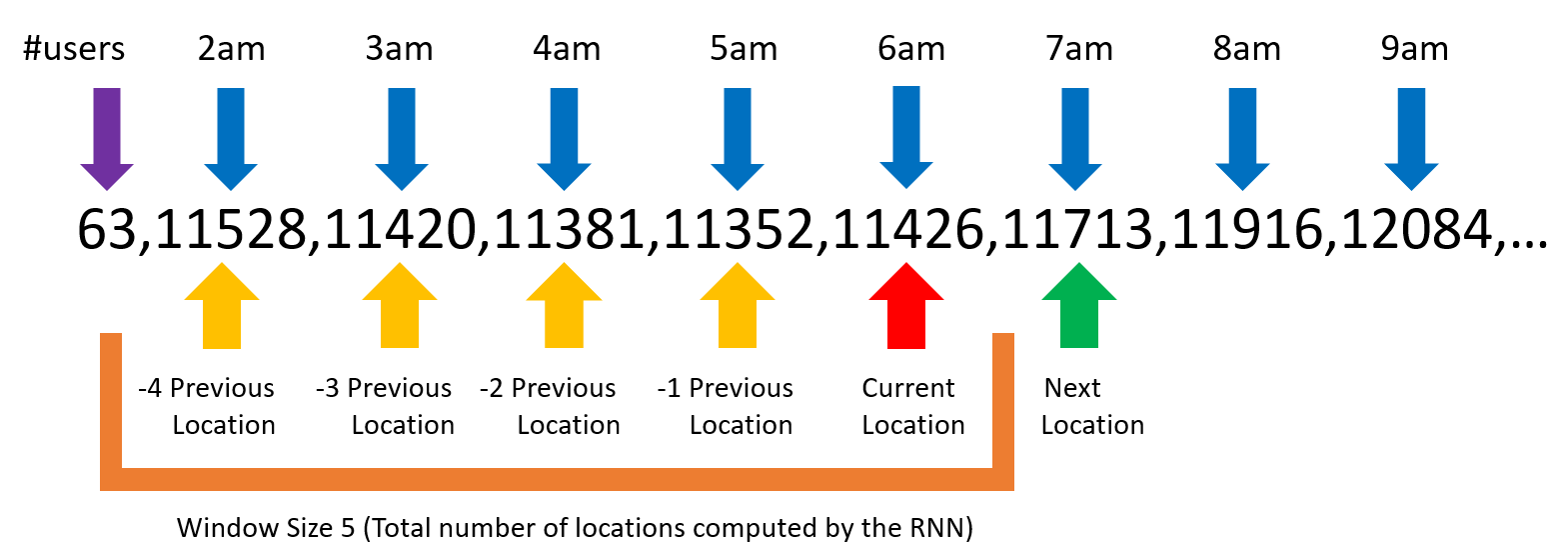}
  \caption{Structure of the dataset}
  \label{figure-dataintro}
\end{figure}


\subsection{Experiment Design}
    \textbf{Handing imbalanced data.} We have used all of the data across all seven days to train, and test our model and optimize every hyperparameter. This includes all $197,415$ unique $24$-hour location paths. We weighed each of the paths equally, removing all class imbalance. This was achieved by ignoring the initial value in the data, which represents the total number of users following the same $24$-hour path, as is illustrated in Figure~\ref{figure-dataintro}. In our provided numerical example, this would be the value of the $63$ users. Doing this both reduces the computational complexity of the prediction problem, as instead of $2,066,000$ $24$-hour location vectors, we have $197,415$; however, it also increases the difficulty of the problem. As every unique $24$-hour location vector is equally valuable to predict correctly in order to maintain a high accuracy, regardless of whether it was a path traveled by $2$ users or $200$ users. Under the original class imballance, we could achieve a high model accuracy by only focusing on the correct prediction of popular paths traveled by a majority of the users. 
    
    \textbf{Split data into training and testing sets.} The $197,415$ unique locations have been randomly split into $157,932$ for training ($80$\%), and $39,483$ for testing ($20$\%). 

    \textbf{Experimental setup and configurations.} With our dataset prepared, we conducted $100$ experiments using an NVIDIA RTX $4090$ graphics processing unit (GPU), and PyTorch version $1.13.1$. These experiments were carried out to study and fine-tune the initial hyperparameters for our work. We initially compared the performance of the Adam Optimizer to Stochastic Gradient Descent (SGD). Afterwards, we proceeded to determine the ideal batch size by scaling in powers of two, starting from a batch size of $16$ and going up to $16,384$. We found a small batch size increased the total training time (for an equal number of epochs) while not fully leveraging the parallel computing capabilities of our GPU. However, very large batches increased the number of epochs necessary to make sufficient model updates (iterations) to reach the same accuracy. After finding an ideal batch size, we determined the necessary number of epochs to fully train a mature model. 

\textbf{Hyperparameter tuning process.} We conducted the hyperparameter tuning process with the aim of ensuring that our RTX $4090$ GPU could fully train the model on the complete dataset within a reasonable amount of time. The tuning process involved conducting experiments to find the optimal value for the Adam Optimizer, which was $0.00024$ for our specific application. We also identified the optimal batch size of approximately $2000$-$2500$. Additionally, the experiments showed that training a fully mature model could be accomplished within $150$ epochs, equivalent to approximately $200,000$ iterations with a batch size of $2500$. This allowed us to train a mature model on our entire dataset in $60$ minutes on average. Thus allowing twenty-four possible hyperparameter combinations to be tested per day.

Optimizing these fundamental hyperparameters required for efficient model training, created our staging point. From this point we could begin performing $100$ experiments to study how the various model hyperparameters influence both the accuracy, and total number of model parameters of our ML model, as discussed in Section \ref{sec:results-discussion}.

\section{Results and Discussion}
\label{sec:results-discussion}

In this section, we present the results to address the research questions, and discuss their practical implications.
    
\begin{table}[t]
\footnotesize   
\centering
\caption{Comparing our model to the baseline.}
\label{table-model-comparison}
\begin{tabular}{l| l| l| l} 
\hline
RQ & Hyperparameter & Our Model & Crivellari Model \\ [0.1ex] 
\hline
  & Architecture & RNN & LSTM \\
  & Num Layers & 2 & 2 \\
RQ1.1  & Embedding Dims & 10 & 100 \\
  & Hidden Nodes & 533 & 4000 \\
  & Window Size & 5 & 4 \\
  & Num Classes & 2134 & 2134 \\
  & Batch Norm & Yes & Yes \\
  & Dropout & No & No \\
  & Optimizer & Adam & Adam \\
  & Loss Function & Cross Entropy & Cross Entropy \\
\hline
  & Accuracy (\%) & 82.54 & 80.16 \\
  & Num Parameters (M) & 2 & 202 \\ 
RQ1.2 & Model Size (MB) & 8 & 791 \\
  & GPU memory (MB) & 317 & 6929 \\
  & Training Time (mins) & 52 & 237 
 \\
\hline
\end{tabular}
\end{table}

\subsection{RQ1. A simpler ML architecture in the context of next
location prediction.}

\textbf{RQ1.1}.  Table~\ref{table-model-comparison} summarized the architectural and hyperparameter differences between our model and Crivellari's work as our baseline \cite{Crivellari2020}. 

As can be seen in the table, we were able to decrease the number of parameters compared to the state-of-the-art work in the domain of next location prediction. Specifically, we were able to decrease the parameter count by a factor of $100$. In addition to the significant decrease in the number of model parameters, we also achieved an additional benefit of a slightly higher accuracy, where we achieved an accuracy of $82.54$ compared to $80.16$ reported in the baseline. 

The improvements to model accuracy can be attributed to several factors. Firstly, careful selection and optimization of hyperparameters was performed, ensuring a plateau of accuracy reached for every given model hyperparameter. Additionally, the simplicity of the model helped mitigate some overfitting issues, enhancing its ability to generalize well to unseen testing data. 

Key hyperparameters such as the window size, number of hidden nodes, and number of embedding dimensions are discussed in Section \ref{se:RQ2}. 

Based on our experiments, LSTMs provide $1.5\%$ higher prediction accuracy compared to RNNs, however they require four times as many model parameters, as well as significantly more training time. Consequently, our $2$ million total parameter RNN architecture would increase to $4.5$ million total parameters just with the switch to an LSTM-based model, which in our view does not justify the $1.5\%$ increase in model accuracy.

\textbf{RQ1.2}. The advantages of a model with a one hundred fold reduction in the number of parameters are numerous and critical for its deployment potential and practical success. As shown in Table~\ref{table-model-comparison}, the benefits include a factor of four reduction in training time on an Nvidia RTX $4090$ GPU, decreased GPU memory requirements by a factor of twenty, allowing the model to be trained on very modest GPus, rather than very powerful and expensive GPUs, and a factor of one hundred reduction in the storage requirements of the model, from $791$ MB down to $8$ MB. 

Additionally, the small model size enables the deployment for inference on modest base stations and edge devices which may have humble energy reserves, and well as power limitations. Even on the latest mobile devices, storing a $791$ MB model represents a significant investment of storage space, whereas an $8$ MB model has much less of a footprint. Additionally, with a smaller model there are benefits such as faster inference speed, as there are far fewer computations to perform, and lower energy consumption during inference, as well as easier deployment and maintainability. Lastly, a model with fewer parameters provides greater interpretability, reduced susceptibility to adversarial attacks, faster experimentation, easier model maintenance, and better performance in transfer learning and online learning \cite{dropout2014}.

\begin{figure}[t]  
  \centering
  \includegraphics[width=0.45\textwidth]{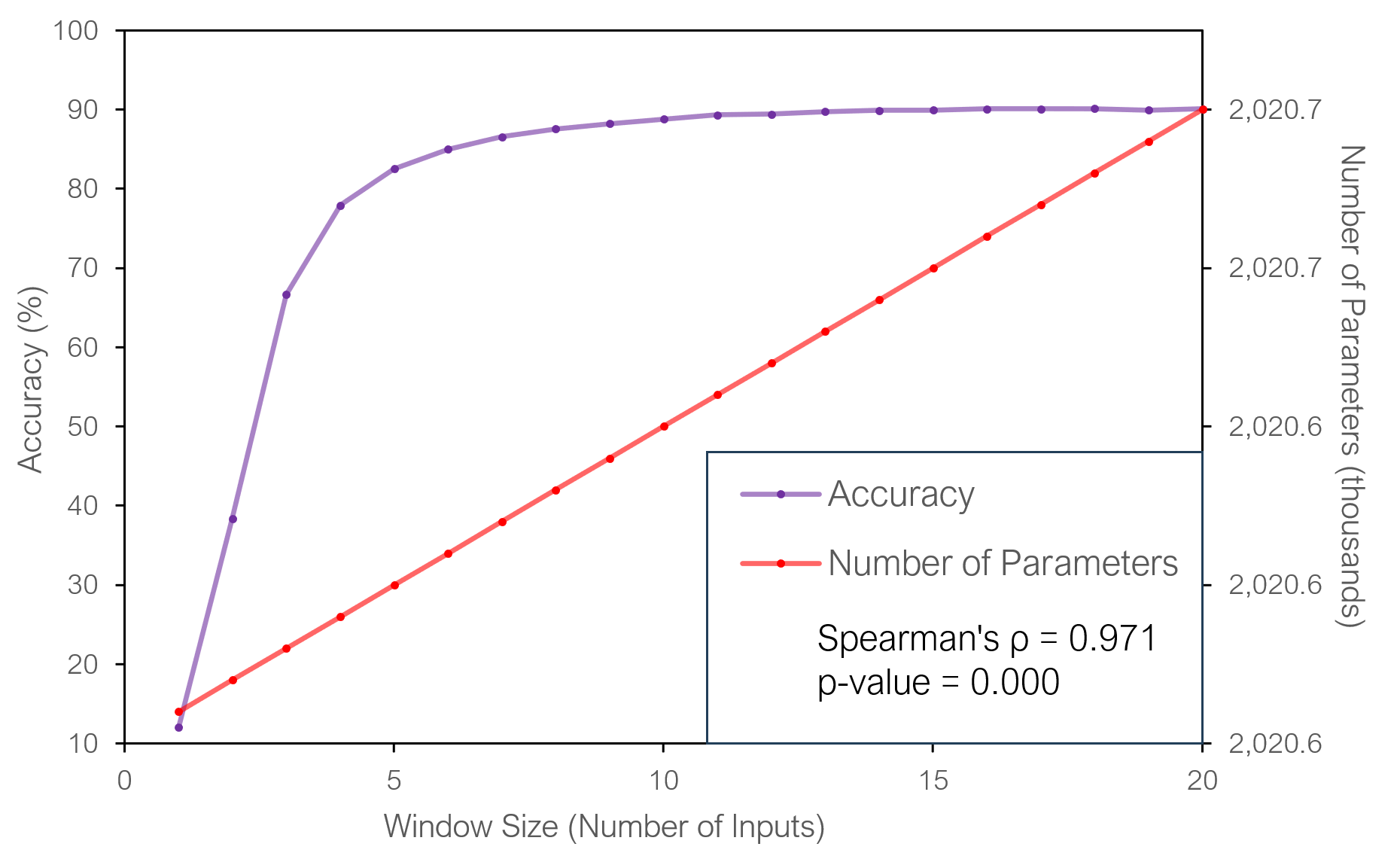}
  \caption{Model accuracy and its relation to window size (total input locations).}
  \label{figure-window}
\end{figure}

\subsection{RQ2. The impact of scaling the model with different hyperparameter values.}\label{se:RQ2}

\textbf{RQ2.1.} As we can see from Figure~\ref{figure-window} there is a clear correlation between expanding the window size (i.e., total number of input locations), and the accuracy achievable by the model. Increasing the window size does not lead to a corresponding increase in the number of model parameters. However, it does come at the expense of the amount of temporal user location information required for the model, and the associated computational time and energy for inference. If those constraints are not an issue for a specific next location prediction application; meaning the computational burden is low, and the wireless carrier has extensive historical location data on their user. For example, if the previous ten locations are available for context, then an accuracy of $89.30\%$ can be achieved. 

As we can see from Figure~\ref{figure-window}, window size continues to improve accuracy up to approximately a size of $10$, meaning the current location of the person, as well as the last $9$ locations. However the diminishing returns begin at approximately a window size of $5$. For our dataset, which starts at 2:00 am, a window size of $5$ enables the very first prediction to be at 7:00 am, which has high utility as that is the beginning of the hours of high daily activity. 

We also used Spearman’s rank correlation to assess the relationship between the window size and  accuracy of the model. Spearman rank correlation is a non-parametric test without any conditions about the data distribution. It is used when the variables are monotonically related and measured on a scale that is at least ordinal. As shown in Figure~\ref{figure-window}, the  window size and the model accuracy are strongly correlated (Spearman’s $\rho$) with $\rho$=$0.971$.

\begin{figure}[t]  
  \centering
  \includegraphics[width=0.45\textwidth]{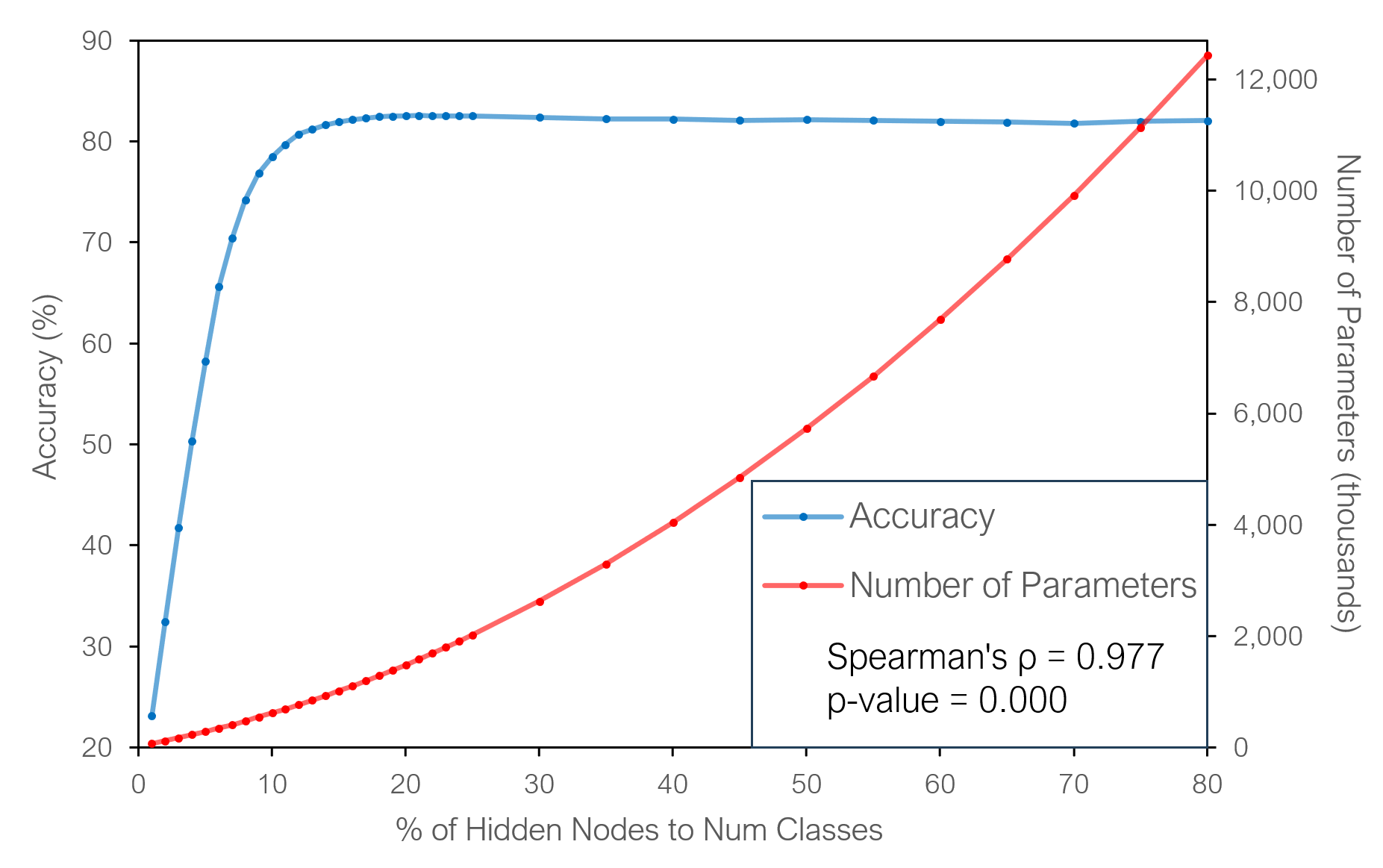}
  \caption{Model accuracy and its relation to the amount of hidden nodes.}
  \label{figure-hidden-node}
\end{figure}

\textbf{RQ2.2.} Figure~\ref{figure-hidden-node} shows that there is a correlation between the number of hidden nodes and the accuracy achievable by the model. To ensure consistency across different dataset sizes and numbers of classes (base stations or locations that a user could potentially visit), the number of hidden nodes is linked to the number of classes.

For our dataset, which consists of $2132$ unique locations, we chose to use $25\%$ of hidden nodes to number of classes, as this maximizes accuracy without being far along on the accuracy plateau. Applying the $0.25$ factor to $2132$ unique locations results in a total of $533$ hidden nodes. Increasing the number of hidden nodes has the largest impact out of all model hyperparameters on both the training time and the number of model parameters. This exponential increase to the model parameters is due to each additional hidden nodes being fully connected with all existing hidden nodes. 

Our statistical analysis also shows that there is a significant correlation (Spearman’s $\rho$) with $\rho$=$0.977$ regarding the number of hidden nodes and the accuracy of the model. For this Spearman analysis we disregarded ''\% hidden nodes to number of classes'' beyond $25\%$, as after this value the hidden nodes plateau and are no longer increasing.

\begin{figure}[t]  
  \centering
  \includegraphics[width=0.45\textwidth]{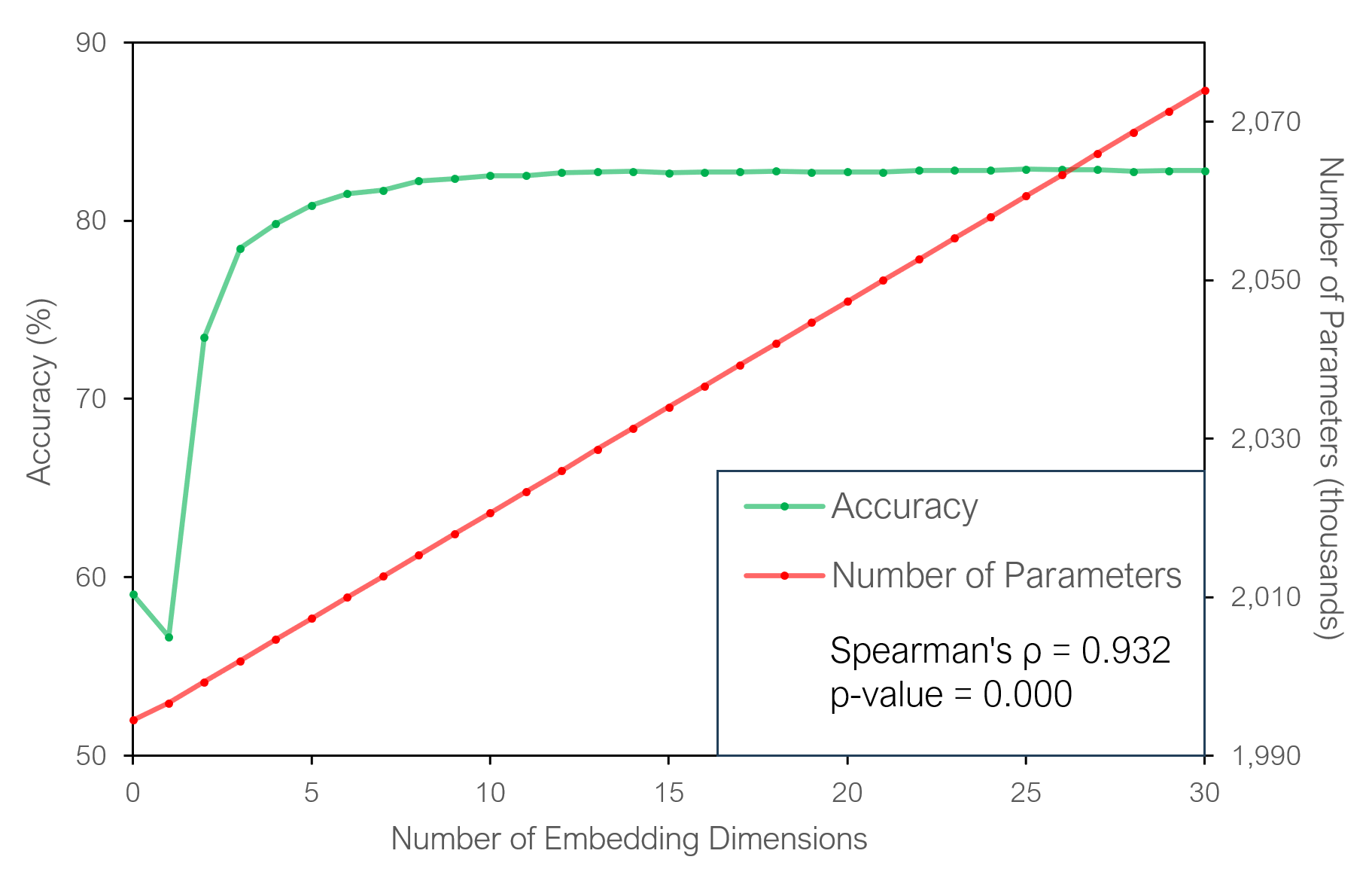}
  \caption{Model accuracy and its relation to the number of embedding dimensions (total learned features per location).}
  \label{figure-embedding}
\end{figure}

\textbf{RQ2.3.} Feature embeddings are a valuable means of encoding the location-based relationships (features) necessary for the ML model to accurately determine the next location a user may choose to visit. Figure~\ref{figure-embedding} shows that the number of embedding dimensions has a very small impact on the overall number of model parameters, but is a crucial aspect of achieving model accuracy (with $\rho$=$0.932$). After $10$ embedding dimensions are provided to the model, a plateau is reached, indicating that introducing additional embedding dimensions provides minimal benefits. That is why we selected this a value of $10$ for this hyperparameter. Interestingly, having no embedding dimensions actually performs slightly better than having a single embedding dimension. This shows that the location ID alone is a superior feature for the model as compared to a single embedding dimension. However, increasing the embedding to just two dimensions performed surprisingly well. 

\begin{figure}[b]  
  \centering
  \includegraphics[width=0.45\textwidth]{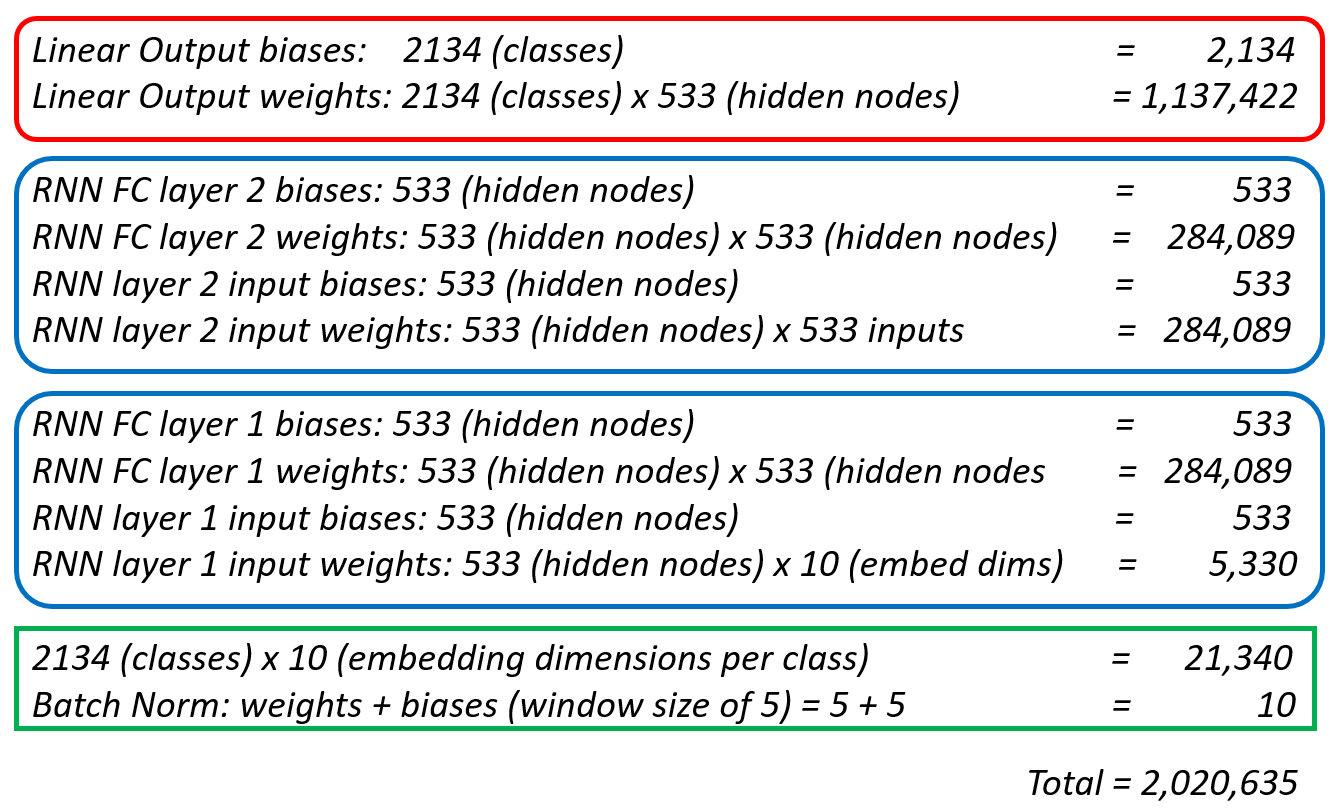}
  \caption{An accounting of all the parameters in our ML model}
  \label{figure-parameters}
\end{figure}

An detailed accounting of every parameter used in our model is shown in Figure~\ref{figure-parameters}. The figure reveals that the majority of all model parameters are in the final linear layer at 1.1 million, demonstrating there is little room for further parameter reduction. Additionally, the figure shows adding a third RNN layer would only increase the total number of parameters by 569 thousand, which could be warranted in some unforeseen complex dataset or application. We can also see the exponential impact that the number of hidden nodes has to the total number of model parameters.

\begin{table}
\centering
\caption{Model and Training Hyperparameters used in our Architecture}
\label{table-scaling-hyperparameters}
\begin{tabular}{c c c} 
 \hline
 Hyperparameter & Equation \\ [0.1ex] 
 \hline
Architecture & RNN \\
Num Layers & 2 \\
Embedding Dims & 10 \\
Hidden Nodes & (Num Classes * 0.25) \\
Window Size & 5 \\
Batch Size & 2500 \\
Num Batches & Total Data / Batch Size \\
Num Iterations & 200,000 \\
Epochs & (200,000 / Num Batches) \\
Optimizer & Adam \\
Learning Rate & 0.00024 \\
Loss Function & Cross Entropy \\
\hline
\end{tabular}
\end{table}

We summarized the model hyperpameters as well as the training hyperparameters we used to train our ML architecture in Table~\ref{table-scaling-hyperparameters}. Through this table, the optimal hyperparameters and also the optimal training regimen can be easily calculated for our optimized ML architecture.

\section{Conclusion}
\label{sec:conclusion}

In this paper, we proposed a simple ML architecture for mobility prediction, while ensuring high accuracy. We have accomplished this objective by conducting one hundred experiments and fine-tuning five seperate architectural hyperparameters. As the results show, we achieved a reduction in the number of model parameters by a factor of $100$ compared to what was reported in the literature. We also slightly increased the accuracy from $80.16\%$ to $82.54\%$. This simplicity makes our model less resource-intensive for both storage and also inference, making it more accessible for a wider range of hardware. One valuable expansion to this work would be to acquire different mobility datasets from different cities, ideally with a different temporal resolution to the one hour we have used in this work. These new datasets could be used to validate or adjust the generalized algorithm presented in this paper to create a ML model on any city-sized dataset. The ultimate goal would be to have an algorithm that would allow an accurate and low parameter ML model to be immediately generated from any city-wide mobility dataset. 

\vspace{0.1cm}

\textbf{ACKNOWLEDGEMENT}
The authors wish to acknowledge the support of Ericsson Canada and the MITACS Accelerate program.

\bibliographystyle{IEEEtran}
\bibliography{main.bib}

\end{document}


\title{Appendix for: ``Scalable and Accurate Test Case Prioritization in Continuous Integration Contexts"}
\maketitle
\begin{appendices}
\section{Metric Definitions}
\label{appendix1}

\bibliographystyle{IEEEtran}
\bibliography{main.bib}